\documentclass{article}
\usepackage[preprint]{corl_2026} % Use this for the initial submission.
\usepackage[numbers]{natbib}
\usepackage{multicol}
\usepackage{graphics} % for pdf, bitmapped graphics files
\usepackage{epsfig} % for postscript graphics files
\usepackage{mathptmx} % assumes new font selection scheme installed
\usepackage{times} % assumes new font selection scheme installed
\usepackage{amsmath} % assumes amsmath package installed
\usepackage{amssymb}  % assumes amsmath package installed
\usepackage{booktabs} % For professional looking tables

\usepackage[utf8]{inputenc} % allow utf-8 input
\usepackage[T1]{fontenc}    % use 8-bit T1 fonts
\usepackage{url}            % simple URL typesetting
\usepackage{multirow}
\usepackage{amsfonts}       % blackboard math symbols
\usepackage{nicefrac}       % compact symbols for 1/2, etc.
\usepackage{microtype}      % microtypography
\usepackage{xcolor}         % colors
\usepackage{graphicx}
\usepackage{wrapfig}
\usepackage{float}
\usepackage{caption}
\usepackage{pifont}
\usepackage{adjustbox}
\usepackage{subcaption}

\usepackage{algorithm}
\usepackage{algorithmic}

\title{CoRMA: Contrastive RMA for Contact-Rich Meta-Adaptation}

\author{
  Wentian Wang\\
  Synthoid AI\\
  \And 
  Chutong Wen\\
  Synthoid AI\\
  \And 
  Hongxu Ma\\
  Synthoid AI\\
  \And 
  Wuhao Wang\\
  Synthoid AI\\
  \And 
  Zhexiong Xue\\
  Synthoid AI\\
  \And 
  Abdul Haseeb Nizamani\\
  Synthoid AI\\
  \And 
  Dandi Zhou\\
  Synthoid AI\\
  \And 
  Xinhai Sun\\
  Synthoid AI\\
  \And 
  Jian-Qiao Zhu\\
  HKU\\
}

\begin{document}
\maketitle

\begin{abstract}
We present \textbf{CoRMA} (Contrastive Robotic Motor Adaptation), a context-based meta-adaptation framework that modifies RMA for force-dominant assembly. 
CoRMA replaces raw simulator-parameter adaptation with a compact 6D simulator-only semantic contact context describing contact onset, lateral engagement, guided transition, contact direction, and jamming. 
A deployable causal Transformer adapter infers this context online from force, proprioceptive, and action histories using semantic regression and a force-regime contrastive objective. 
At deployment, oracle context is removed and replaced by the inferred context, enabling within-episode adaptation without demonstrations, privileged inputs, or gradient updates. 
We evaluate CoRMA on PegInsert, GearMesh, and NutThread in Isaac Lab / Isaac Sim 5.0 and deploy on a real Marvin arm. 
Compared with FORGE baselines that achieve high simulation success but degrade substantially on hardware, CoRMA retains higher verified real success under controlled target-pose noise. 
These results support semantic contact inference as a reusable adaptation interface within a related assembly task family, while broader unseen-task generalization and Real2Sim calibration remain future work.
\end{abstract}

\keywords{Contact-Rich Assembly, Sim-to-Real, Meta Learning} 

\begin{figure*}[t]
  \centering
  \includegraphics[width=\textwidth]{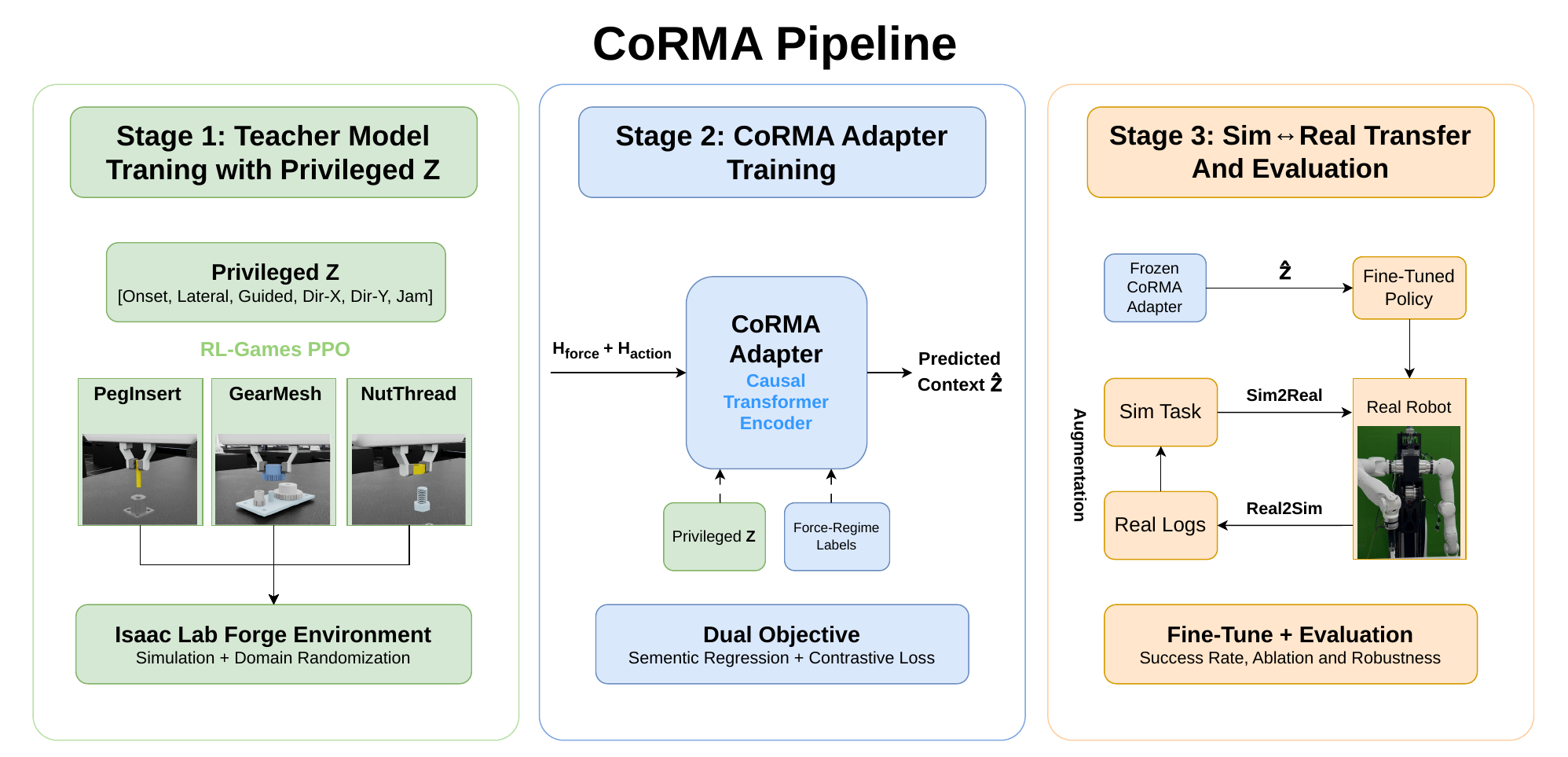}
  \caption{\textbf{CoRMA pipeline overview.}
Stage 1 trains privileged contact-aware policies in Isaac Lab FORGE using RL-Games PPO with a simulator-only 6D semantic contact latent $z_t$.
Stage 2 trains a causal Transformer adapter to infer $\hat{z}_t$ from deployable force, proprioceptive, and action histories, using semantic regression and force-regime contrastive learning.
Stage 3 removes oracle $z_t$ and injects the adapter prediction $\hat{z}_t$ into the policy during fine-tuning and real deployment.
This implements an RMA-style non-cheating deployment interface for context-based meta-adaptation: privileged contact semantics are used only as simulation supervision, while the deployed controller adapts online from sensor history.
Real logs are used for deployment diagnostics and motivate future Real2Sim calibration.
}
  \label{fig:meta_pipeline}
\end{figure*}

\section{Introduction}

Contact-rich robotic assembly~\citep{factory,industry,AutoMate} remains a long-standing challenge in robotics. 
Tasks such as PegInsert, GearMesh, and NutThread require precise geometric alignment, sustained contact regulation, and recovery from intermittent failures over long horizons. 
Unlike free-space manipulation, small mismatches in contact geometry, friction, compliance, sensing, or actuation can produce jamming, unstable transitions, or irreversible failure. 
Robust assembly therefore depends not only on generating motion, but also on interpreting what a sequence of contact forces \emph{means} during execution.

Recent sim-to-real assembly systems, including FORGE~\citep{forge}, show that reinforcement learning can produce robust contact-rich behaviors under uncertainty. 
However, these policies are still commonly trained and tuned in a task-specific manner. 
New parts, fixtures, or objectives may require new policy training or additional real-world calibration, even when the underlying contact mechanisms are closely related. 
Many successful manipulation systems also rely on demonstration-heavy pipelines, such as behavior cloning~\citep{Pomerleau,dagger}, imitation learning~\citep{SCHAAL,abbeel,gcl,gail}, or residual sim-to-real fine-tuning~\citep{res,respo,resbc}. 
Classical robotics has addressed manipulation uncertainty through adaptive control and disturbance compensation, typically for tracking and stabilization rather than for learning reusable cross-task contact representations~\citep{Tony}. 
While effective, these approaches can be costly to extend across task variants and operating conditions because they require repeated data collection, calibration, or adaptation whenever the setup changes~\citep{data}.

This motivates a meta-adaptation view of contact-rich assembly~\citep{meta,maml}: instead of learning each assembly policy in isolation, the robot should infer an interaction context online and reuse this context across related tasks. 
Teacher--student frameworks such as Rapid Motor Adaptation (RMA)~\citep{rma} provide a useful template for this goal. 
RMA trains a privileged teacher in simulation and learns an adapter that predicts privileged context from deployable sensor history, so privileged information is used only during training while deployment relies only on onboard observations. 
For contact-rich assembly, however, the standard RMA formulation is not directly sufficient. 
Many manipulation-oriented adaptation systems rely on visual or object-centric cues, whereas our deployment setting uses force, proprioception, and action history. 
Once contact begins, the critical question is not only where the object is, but whether the interaction is in free motion, first touch, lateral engagement, guided sliding, thread or tooth engagement, or a jammed regime.

We propose CoRMA, a force-based context adaptation framework that modifies RMA for contact-rich assembly. 
CoRMA is built on the hypothesis that related assembly tasks share recurring \emph{semantic contact structure}. 
Instead of using raw simulator parameters as privileged context, CoRMA constructs a compact 6D \textbf{Privileged Z} describing contact onset, lateral engagement, guided contact transition, contact direction along two axes, and jam or stick--slip tendency. 
Force is treated as deployable evidence, while Privileged Z represents the semantic interpretation of contact that is useful for control. 
Because raw force histories can be ambiguous under pointwise regression, CoRMA uses a causal Transformer adapter to infer $\hat{z}_t$ from force/action histories and adds a force-regime contrastive objective~\citep{single,crl} to organize histories by contact semantics rather than by raw trajectory similarity.

Fig.~\ref{fig:meta_pipeline} summarizes the CoRMA pipeline. 
Stage~1 trains privileged contact-aware policies in Isaac Lab FORGE using RL-Games PPO with $z_t$ appended to the observation. 
Stage~2 trains the CoRMA adapter to predict $\hat{z}_t$ from deployable history using semantic regression and force-regime contrastive learning. 
Stage~3 removes oracle $z_t$ and injects the adapter prediction $\hat{z}_t$ into the policy during fine-tuning and real deployment. 
In this sense, CoRMA is a context-based meta-adaptation system rather than a MAML-style gradient inner-loop method: adaptation happens online through the inferred contact context, without demonstrations, privileged inputs, or test-time gradient updates. 
We evaluate CoRMA on PegInsert, GearMesh, and NutThread in simulation and on a real Marvin arm. 
Our current evaluation tests reuse within a related contact-rich assembly family rather than unconstrained generalization to arbitrary unseen tasks, and the results show that semantic contact inference can reduce the real-world degradation observed in strong task-specific FORGE baselines.

\section{Related Works}

\subsection{Meta-Reinforcement Learning and Context Adaptation}

Meta-RL methods aim to train agents that can adapt rapidly to new tasks or dynamics by leveraging experience over a task distribution. 
Gradient-based methods such as MAML~\citep{maml} learn parameters that can be fine-tuned with a small number of gradient steps, but require inner-loop updates and additional adaptation data at test time. 
Recurrent meta-RL methods such as RL$^2$~\citep{rl2} instead learn an implicit adaptation procedure in the hidden state of a recurrent policy. 
Context-based methods such as PEARL~\citep{pearl} and variBAD~\citep{vari} infer an explicit latent context under partial observability, allowing the policy to condition on an estimated task or dynamics belief.

CoRMA follows this context-based view, but specializes it to contact-rich assembly. 
Rather than learning a generic task embedding only from rewards, CoRMA uses simulator-derived semantic contact labels as supervised targets for an amortized context estimator. 
At deployment, adaptation occurs through online latent inference from force, proprioceptive, and action history, without demonstrations or MAML-style gradient updates. 
Thus, CoRMA is best understood as context-based meta-adaptation within a related assembly task family.

\subsection{Privileged Adaptation and Rapid Motor Adaptation}

RMA trains a privileged teacher policy in simulation and then learns an adaptation module that predicts the teacher's latent context from deployable history. 
This privileged-train / deployable-test paradigm is attractive for robotics because simulator-only quantities can guide training while the deployed policy remains non-cheating. 
Manipulation-oriented RMA variants extend this idea to object manipulation, often using visual or object-centric information to infer privileged properties~\citep{rma4rma}.

CoRMA keeps the RMA deployment principle but changes three components for force-dominant assembly. 
First, the privileged target is not a broad vector of raw simulator parameters or visual object properties, but a compact semantic contact context describing contact onset, lateral engagement, guided transition, directional bias, and jamming tendency. 
Second, the adapter uses deployable force/proprioceptive/action history rather than camera input. 
Third, the history encoder and objective are modified: CoRMA uses a causal Transformer and a force-regime contrastive loss to structure contact histories by interaction semantics.

\subsection{Long-Horizon History Encoders and Contrastive Representation Learning}

Self-attention provides a mechanism for aggregating sparse evidence over long horizons. 
In reinforcement learning, Transformer-based architectures such as GTrXL~\citep{stable} have been used to improve memory and stability in partially observable sequential decision problems. 
This is relevant for contact-rich assembly because brief force transients, early contact onset, or delayed jam signatures can affect later corrective actions.

Contrastive learning has been widely used to improve representation quality in RL and robotics, including pixel-based methods such as CURL~\citep{curl} and general contrastive objectives such as InfoNCE~\citep{single,crl}. 
CoRMA uses contrastive learning not as a visual pretraining objective, but as a weak semantic regularizer for contact-context inference. 
The adapter retains a supervised regression head for the 6D contact context, while an auxiliary contrastive head groups histories with similar force-regime labels, such as first contact or guided sliding. 
This separates numerical context prediction from representation structuring and avoids replacing the privileged regression objective.

\begin{figure*}[t]
  \centering
  \includegraphics[width=\textwidth]{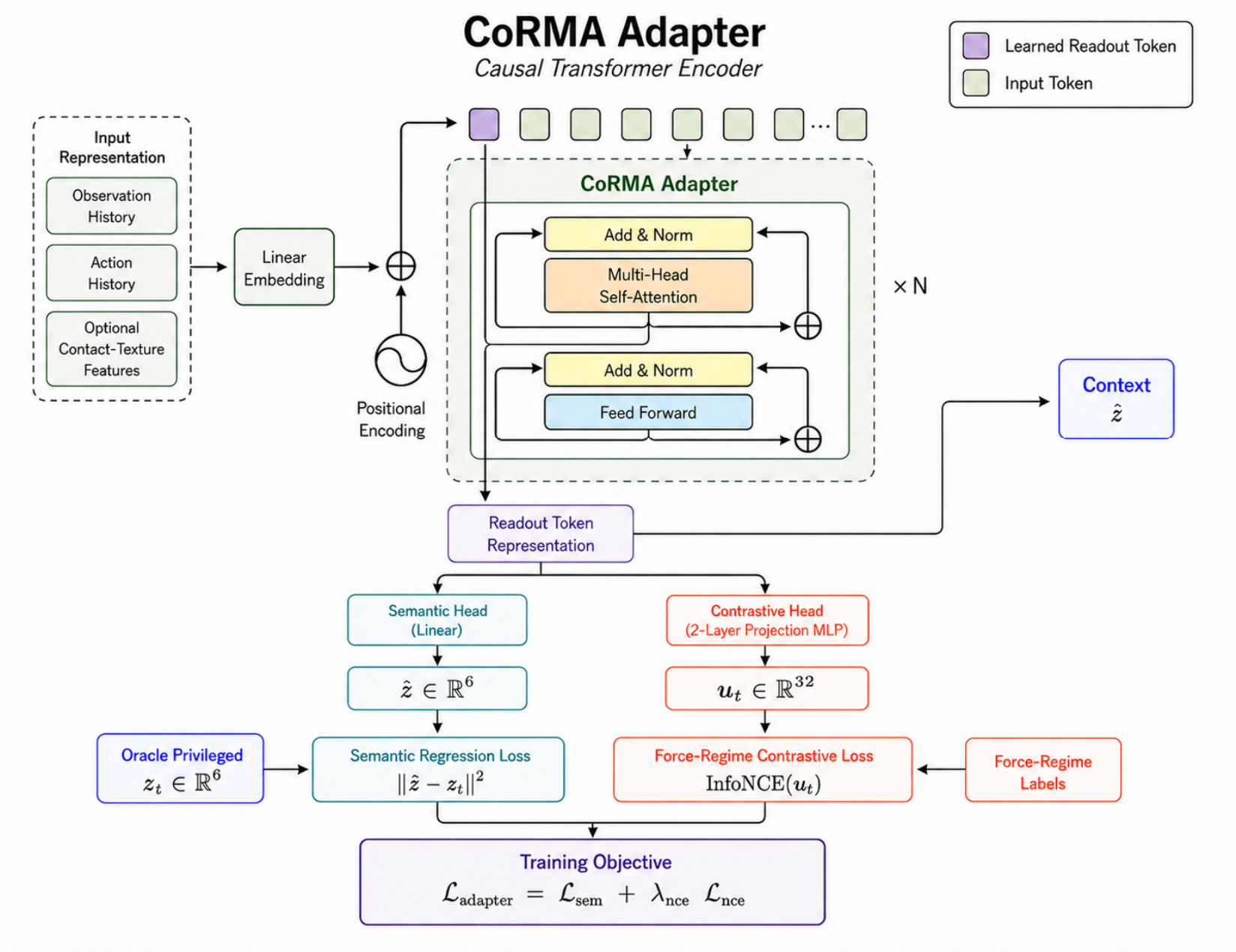}
  \caption{\textbf{CoRMA adapter.}
  A causal Transformer encodes deployable force/proprioceptive/action history with a learned readout token.
  The semantic head predicts the 6D context $\hat z$ used by the policy, while an auxiliary contrastive head produces $u_t$ for force-regime InfoNCE.
  Positives are defined by coarse contact regimes: free motion, first contact, guided sliding, and jamming.
  }
  \label{fig:adapter}
\end{figure*}

\section{Method}
\label{sec:method}

CoRMA formulates contact-rich assembly as a partially observable context-adaptation problem. 
At timestep $t$, the robot observes deployable signals $o_t$, including proprioception, force/torque measurements, force-threshold information, and previous actions, but it does not directly observe the semantic contact state that explains the interaction. 
CoRMA therefore learns a deployable adapter that infers a compact contact context from recent history and injects this context into the policy. 
Unlike visual RMA variants, CoRMA uses no camera input at deployment; the evidence for adaptation comes from force, proprioceptive, and action histories.

\subsection{Semantic Contact Context}

Instead of adapting through a broad vector of raw simulator parameters, CoRMA defines a simulator-only semantic latent, Privileged $Z$. 
In the current implementation, $z_t=[z_t^{\mathrm{onset}},z_t^{\mathrm{lateral}},z_t^{\mathrm{guided}},z_t^{\mathrm{dir}\text{-}x},z_t^{\mathrm{dir}\text{-}y},z_t^{\mathrm{jam}}]\in\mathbb{R}^{6}$, where the dimensions represent contact onset, lateral engagement, guided contact transition, contact direction along two axes, and jam or stick--slip tendency. 
These scores are computed from simulator contact and force dynamics and are used only during training. 
The distinction is important: force is deployable evidence, while $z_t$ is the semantic interpretation of that evidence. 
This gives PegInsert, GearMesh, and NutThread a shared contact vocabulary even though their geometries, rewards, and success checks differ.

\subsection{Privileged Teacher Rollouts}

For each task in the evaluated assembly family, we train a privileged teacher policy in Isaac Lab~\cite{isaaclab} using RL-Games PPO~\cite{ppo} with observations augmented by Privileged $Z$, written compactly as $a_t\sim\pi_i(a_t\mid o_t,z_t)$, where $i$ indexes the task. 
The teacher is not directly deployable because $z_t$ is simulator-only. 
Its role is to generate contact-rich trajectories and provide supervised semantic-context targets for adapter training. 
Teacher rollouts are converted into history-window samples $\mathcal{D}=\{(o_{t-H+1:t},a_{t-H+1:t},z_t,c_t)\}$, where $H$ is the history length and $c_t$ is a weak force-regime label. 
Sampling windows across the entire trajectory exposes the adapter to approach, first contact, guided motion, and jam/failure phases rather than only terminal outcomes.

\subsection{CoRMA Adapter and Contrastive Objective}

Fig.~\ref{fig:adapter} shows the CoRMA adapter. 
The adapter maps deployable history to a semantic context prediction $\hat z_t=\phi(o_{t-H+1:t},a_{t-H+1:t})$. 
A causal Transformer with a learned readout token encodes the history into $h_t=f_\theta(o_{t-H+1:t},a_{t-H+1:t})$. 
The readout representation is passed to two heads: a semantic head $\hat z_t=g_{\mathrm{sem}}(h_t)$ and an auxiliary contrastive head $u_t=g_{\mathrm{nce}}(h_t)$. 
Only $\hat z_t$ is used by the policy; $u_t$ is used only during adapter training to shape the representation.

The semantic head is trained to regress all six Privileged $Z$ dimensions with $\mathcal{L}_{\mathrm{sem}}=\|\hat z_t-z_t\|_2^2$. 
We do not use a static/dynamic split: contact onset, lateral engagement, guided transition, contact direction, and jam tendency are all active targets. 
However, pointwise regression alone does not explicitly teach the adapter that histories from different tasks can share the same contact meaning. 
For this reason, CoRMA adds a force-regime InfoNCE objective over the auxiliary embedding $u_t$.

The contrastive labels are weak semantic regimes computed from deployable force evidence: \texttt{free}, \texttt{first\_contact}, \texttt{guided\_slide}, and \texttt{jam}. 
For an anchor history, positives are histories with the same force-regime label and negatives are histories with different labels, regardless of task identity. 
The resulting loss is
\[
\mathcal{L}_{\mathrm{nce}}
=
-\log
\frac{\exp(\mathrm{sim}(u_t,u^+)/\tau)}
{\exp(\mathrm{sim}(u_t,u^+)/\tau)+
\sum_{u^-\in\mathcal{N}_t}\exp(\mathrm{sim}(u_t,u^-)/\tau)},
\]
where $\mathrm{sim}(\cdot,\cdot)$ is cosine similarity and $\tau$ is the temperature. 
This objective encourages histories with similar contact semantics to cluster in the auxiliary space, while the semantic head preserves numerical fidelity to the 6D contact context. 
The Stage~2 objective is $\mathcal{L}_{\mathrm{adapter}}=\mathcal{L}_{\mathrm{sem}}+\lambda_{\mathrm{nce}}\mathcal{L}_{\mathrm{nce}}$. 
We use the contrastive term as a weak semantic regularizer rather than as a replacement for supervised Privileged $Z$ regression.

\subsection{Latent Injection for Deployment}

During fine-tuning and real deployment, oracle Privileged $Z$ is removed. 
The frozen adapter predicts $\hat z_t$ online from deployable history, and the policy receives $\hat z_t$ in place of simulator-computed $z_t$, i.e., $a_t\sim\pi(a_t\mid o_t,\hat z_t)$. 
This preserves the RMA non-cheating principle: privileged contact semantics supervise training, but deployment uses only onboard force/proprioceptive/action history. 
In our evaluation, the same semantic latent-injection interface is used across PegInsert, GearMesh, and NutThread; broader held-out task generalization is left for future work.

\begin{figure}[t]
    \centering
    \begin{subfigure}[t]{0.31\linewidth}
        \centering
        % PegInsert: crop both bottom and top black bars
        \includegraphics[
            width=\linewidth,
            height=0.20\textheight,
            keepaspectratio,
            trim=0 85 0 95,
            clip
        ]{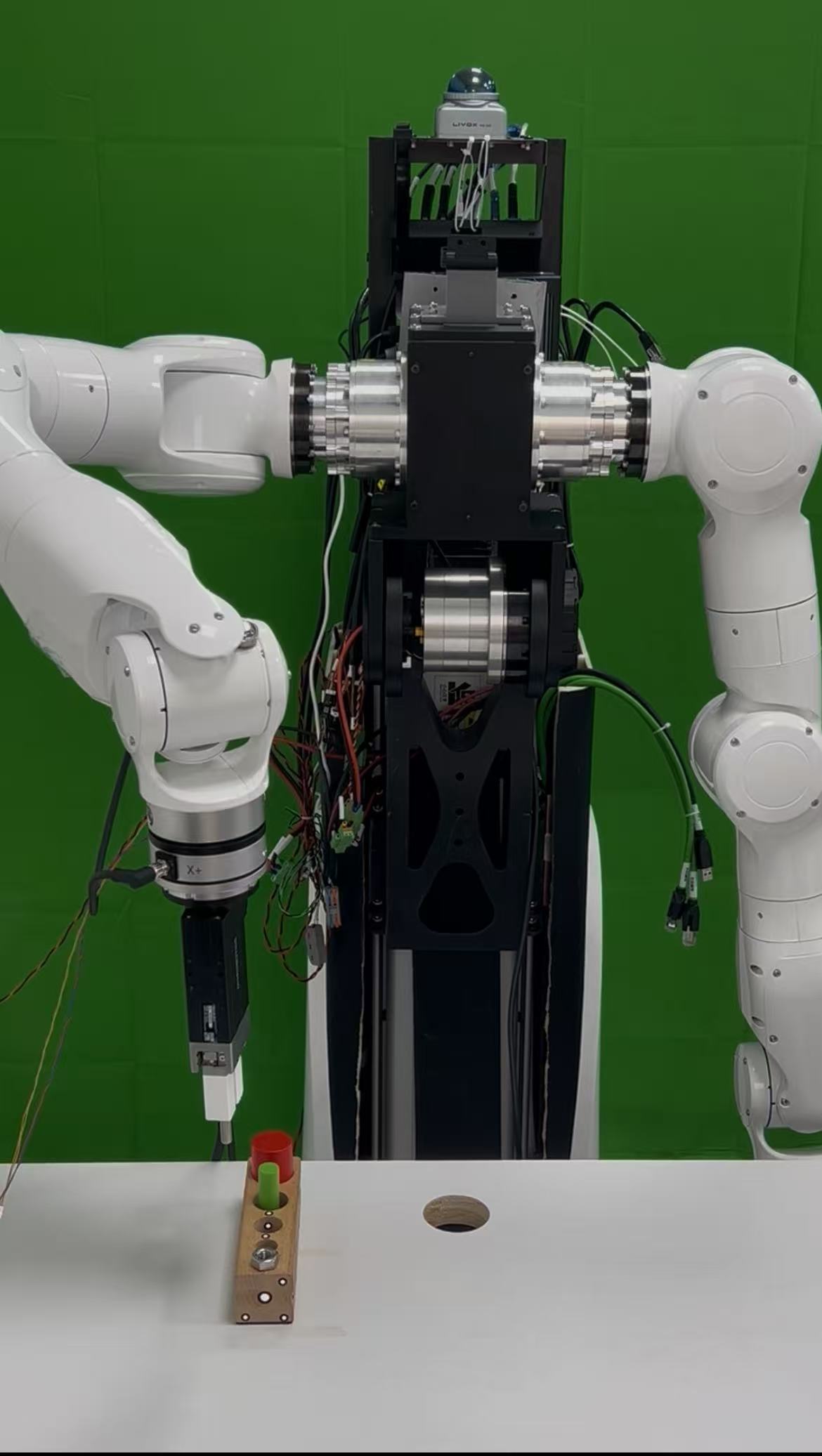}
        \caption{PegInsert}
        \label{fig:real_peg}
    \end{subfigure}
    \hfill
    \begin{subfigure}[t]{0.31\linewidth}
        \centering
        % GearMesh: crop both bottom and top black bars
        \includegraphics[
            width=\linewidth,
            height=0.20\textheight,
            keepaspectratio,
            trim=0 85 0 95,
            clip
        ]{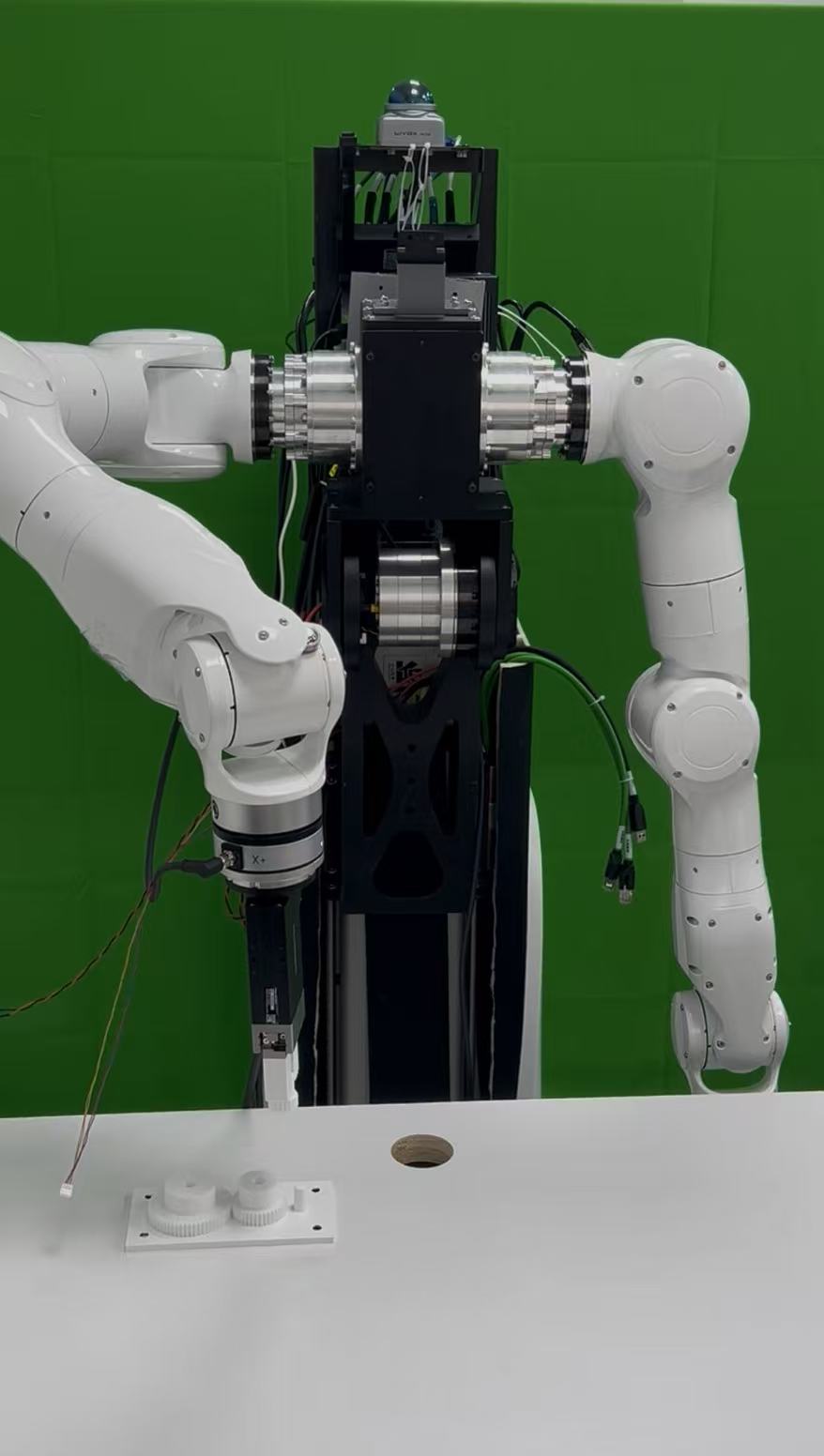}
        \caption{GearMesh}
        \label{fig:real_gear}
    \end{subfigure}
    \begin{subfigure}[t]{0.31\linewidth}
        \centering
        % trim = left bottom right top
        % NutThread: only crop bottom black bar
        \includegraphics[
            width=\linewidth,
            height=0.20\textheight,
            keepaspectratio,
            trim=0 85 0 99,
            clip
        ]{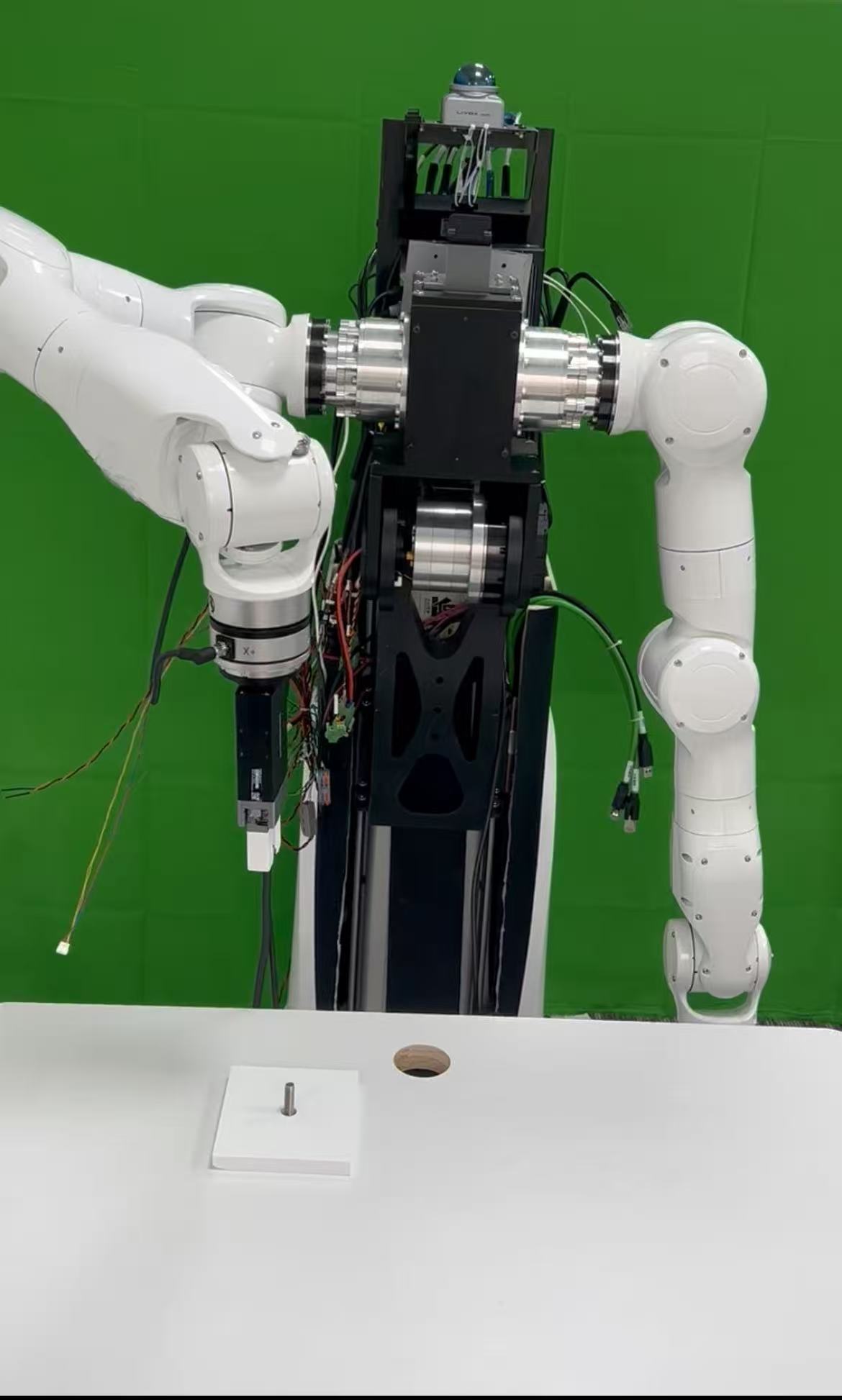}
        \caption{NutThread}
        \label{fig:real_nut}
    \end{subfigure} 
    \caption{
    \textbf{Real-robot deployment tasks.}
    CoRMA and FORGE are evaluated on the same Marvin hardware interface across PegInsert, GearMesh and NutThread.
    }
    \label{fig:real_robot_tasks}
\end{figure}

\section{Experiments}
\label{sec:experiments}

\subsection{Experimental Setup}

We evaluate CoRMA in Isaac Lab / Isaac Sim~5.0 and on the real Marvin arm. 
Our simulator is based on the Isaac Lab FORGE direct RL environment for contact-rich assembly. 
FORGE was originally designed around a 6-DoF Franka/Panda-style task-space action interface; to deploy on our 7-DoF Marvin arm, we preserve the FORGE-style observation, reward, reset, and task definitions as much as possible, and modify only the robot-specific control interface. 
The policy receives deployable proprioceptive, force/torque, force-threshold, and previous-action observations. 
For CoRMA teacher training, we append the simulator-only 6D Privileged $Z$; for FORGE baselines, this latent is removed. 
For real deployment, Cartesian policy targets are converted to Marvin joint commands through TRAC-IK~\citep{tracik}, matching the execution path used by the real robot.

As shown in Fig.~\ref{fig:real_robot_tasks}, we evaluate three FORGE-style assembly tasks: PegInsert, GearMesh, and NutThread. 
PegInsert requires inserting a held bolt into a fixed hole under lateral pose uncertainty. 
GearMesh requires inserting and meshing a gear with discrete tooth engagement. 
NutThread requires long-horizon alignment and threading under sustained contact. 
All three tasks share the same policy observation convention and action interface, while task-specific configuration files define object geometry, reset distributions, reward terms, and success conditions.

The real-robot experiments use approximately 3~mm target-position noise to emulate camera-based localization error. 
Our current prototype does not include an onboard camera, so the noisy target is injected directly into the deployment pipeline. 
This evaluates force/proprioceptive adaptation under controlled pose perturbation, not a full vision-force system. 
Both CoRMA and FORGE are deployed through the same Marvin hardware interface, so the comparison uses the same IK layer, execution stack, force sensor, and verification rule.

\subsection{Training, Baselines, and Evaluation Protocol}

All simulation policies are trained with the Isaac Lab / RL-Games PPO pipeline. 
Stage~1 trains privileged CoRMA teachers with observations augmented by Privileged $Z$. 
Stage~2 collects teacher rollouts and trains the CoRMA adapter from fixed-length histories of force, proprioception, and actions. 
Stage~3 freezes the adapter and fine-tunes or evaluates policies by replacing oracle $z_t$ with the online adapter prediction $\hat z_t$.

The main baseline is FORGE trained under the same task family and deployed through the same Marvin real-robot interface, but without Privileged $Z$ and without an adapter. 
CoRMA is evaluated in student mode, where the frozen adapter predicts $\hat z_t$ online and no oracle Privileged $Z$ is used. 
Each simulation result is computed over 80 episodes. 
For real-robot evaluation, success is computed per logged run by grouping rows by \texttt{run\_id} and taking the maximum \texttt{insertion\_verified} value within each run. 
Runs without a final verified insertion are counted as non-success. 
For FORGE NutThread, all 20 real trials failed the same deployment-side insertion-verification criterion used for CoRMA, so we report the verified success rate as 0/20 rather than treating it as missing data.

\subsection{Simulation and Real-Robot Results}

Table~\ref{tab:main_results} summarizes simulation success, real-robot verified success, sim-to-real degradation, and Wilson 95\% confidence intervals for the real success rates. 
Real success is measured by the final \texttt{insertion\_verified} signal after policy execution and the post-policy probing procedure. 
The confidence intervals are computed from the existing real-robot success/failure counts and quantify finite-sample uncertainty; they do not require additional robot trials.

The key observation is that high simulation success alone is not a reliable predictor of real deployment performance. 
FORGE achieves near-perfect simulation success on GearMesh and PegInsert, but its real verified success drops sharply to 25.0\% and 12.5\%, respectively. 
CoRMA is not uniformly better in simulation: it is slightly lower than FORGE on GearMesh and substantially lower on PegInsert. 
However, under the same Marvin deployment interface, CoRMA retains higher verified real success on both tasks, achieving 65.0\% on GearMesh and 50.0\% on PegInsert. 
The Wilson intervals are wide because the real-robot evaluation contains tens rather than hundreds of trials, but the intervals make this uncertainty explicit.

On NutThread, CoRMA obtains 16/27 verified real successes, while FORGE obtains 0/20 under the same final insertion-verification rule. 
We interpret this comparison cautiously because threading is the task most affected by real contact, actuation, force sensing, and verification mismatch. 
Still, across the evaluated tasks, CoRMA shows a smaller sim-to-real degradation than FORGE. 
These results support the main empirical claim that semantic contact inference can reduce real-world degradation for force-dominant assembly, while also showing that CoRMA does not eliminate the sim-to-real gap.

\begin{table}[t]
\centering
\small
\setlength{\tabcolsep}{5pt}
\caption{
\textbf{Simulation and real-robot verified success.}
Real success is measured by the final \texttt{insertion\_verified} signal.
The gap is real verified success rate minus simulation success rate in percentage points.
Wilson 95\% confidence intervals quantify finite-sample uncertainty in the real-robot success rate.
}
\vspace{4px}
\label{tab:main_results}
\begin{tabular}{llcccc}
\toprule
\textbf{Task} & \textbf{Method} & \textbf{Sim Success} & \textbf{Real Success} & \textbf{Gap} & \textbf{Real 95\% CI} \\
\midrule
\multirow{2}{*}{PegInsert}
& CoRMA & 48/80 (60.00\%) & \textbf{11/22 (50.0\%)} & \textbf{-10.00} & \textbf{[30.7, 69.3]\%} \\
& FORGE & \textbf{79/80 (98.75\%)} & 3/24 (12.5\%) & -86.25 & [4.3, 31.0]\% \\
\midrule
\multirow{2}{*}{GearMesh}
& CoRMA & 74/80 (92.50\%) & \textbf{13/20 (65.0\%)} & \textbf{-27.50} & \textbf{[43.3, 81.9]\%} \\
& FORGE & \textbf{80/80 (100.00\%)} & 5/20 (25.0\%) & -75.00 & [11.2, 46.9]\% \\
\midrule
\multirow{2}{*}{NutThread}
& CoRMA & \textbf{73/80 (91.25\%)} & \textbf{16/27 (59.3\%)} & \textbf{-31.95} & \textbf{[40.7, 75.5]\%} \\
& FORGE & 54/80 (67.50\%) & 0/20 (0.0\%) & -67.50 & [0.0, 16.1]\% \\
\bottomrule
\end{tabular}
\vspace{-0.5em}
\end{table}

\subsection{Adapter and Representation Analysis}

Table~\ref{tab:adapter_summary} summarizes the Stage~2 adapter validation results; full ablations are reported in Appendix~\ref{app:ablation_experiments}. 
The causal Transformer is the dominant architectural factor: replacing it with an RMA-style Conv1D adapter reduces mean validation $R^2$ from 0.8792 to 0.4336 and increases mean MSE from 0.1208 to 0.5664. 
The force-regime contrastive objective gives a smaller but consistent gain over Transformer-MSE, improving mean $R^2$ from 0.8688 to 0.8792 and reducing mean MSE from 0.1312 to 0.1208. 
We therefore interpret InfoNCE as a semantic structuring regularizer rather than the sole driver of performance.

Appendix~\ref{app:stage2_diagnostics} provides complementary diagnostics for the adapter representation. 
The PCA visualizations compare the supervised semantic output $\hat z_t$ with the auxiliary contrastive embedding $u_t$: $\hat z_t$ is optimized to predict the 6D Privileged $Z$ used by the policy, while $u_t$ is optimized only through force-regime InfoNCE. 
Across task pairs, the contrastive space shows clearer organization by coarse contact regimes than the regression output alone, supporting the intended role of InfoNCE as a representation-structuring objective. 
The appendix also reports $\hat z_t$ versus oracle $z_t$ scatter plots and force-regime separability probes, showing that the deployed context remains predictive of the simulator-derived semantic contact labels and retains regime-relevant information. 
These diagnostics support the two-head adapter design, but we do not claim that the learned embedding is fully task-invariant.

On the real robot, oracle Privileged $Z$ is unavailable, so real-world adapter accuracy cannot be measured directly. 
Instead, we use deployment logs diagnostically by comparing predicted $\hat z_t$ with compensated wrench, force onset, end-effector motion, and final insertion verification. 
Appendix~\ref{app:real_force_regimes} shows real wrench-derived patterns motivating the coarse force-regime labels used for contrastive training. 
These diagnostics support the plausibility of the contact-regime labels, but do not replace a full Real2Sim calibration study.

\begin{table}[t]
\centering
\caption{
\textbf{Stage-2 adapter validation summary.}
Mean validation metrics across task pairs. 
The causal Transformer provides the dominant gain over an RMA-style Conv1D encoder, while force-regime InfoNCE gives a smaller regularization benefit over Transformer-MSE.
}
\vspace{4px}
\label{tab:adapter_summary}
\begin{tabular}{lcccc}
\toprule
\textbf{Adapter} & $\mathbf{R^2}$ & \textbf{Pearson} & \textbf{Cos-Sim} & \textbf{MSE} \\
\midrule
RMA-Conv & 0.4336 & 0.6411 & 0.6307 & 0.5664 \\
RMA-Transformer & 0.8688 & 0.9313 & 0.9151 & 0.1312 \\
CoRMA (Ours) & \textbf{0.8792} & \textbf{0.9369} & \textbf{0.9215} & \textbf{0.1208} \\
\bottomrule
\end{tabular}
\end{table}

\section{Limitations}
\label{sec:limitations}

This work has several limitations. 
First, the evaluation covers reuse within three related assembly tasks, but does not include a held-out unseen task; therefore, our results support task-family-level meta-adaptation rather than unconstrained generalization. 
Second, our real deployment uses a 7-DoF Marvin arm rather than the Franka/Panda-style setup used by FORGE, and Cartesian targets are executed through TRAC-IK. 
This introduces additional variables, including IK convergence, joint-limit handling, tracking error, servo timing, and arm-specific compliance, so real success depends on both the learned policy and the deployment stack. 
Third, we have not yet performed Real2Sim calibration from real logs; the remaining sim-to-real gap likely reflects mismatch in dynamics, contact compliance, friction, sensing, latency, and controller timing. 
Fourth, the current force pipeline uses basic smoothing but no dedicated state estimator, so future work should incorporate principled filtering to reduce sensor noise while preserving contact-onset events. 
Finally, the real-robot study contains tens of trials per task rather than a large-scale reliability evaluation, so the reported results should be interpreted as controlled deployment evidence rather than full reliability certification.

\section{Conclusion}
\label{sec:conclusion}

We presented \textbf{CoRMA}, a context-based meta-adaptation framework for contact-rich robotic assembly. 
CoRMA modifies RMA by replacing raw simulator-parameter adaptation with a compact semantic contact context, using a causal Transformer adapter to infer this context from force/proprioceptive/action history, and adding a force-regime contrastive objective for semantic representation structuring. 
At deployment, oracle Privileged $Z$ is replaced by the adapter prediction $\hat z_t$, enabling online contact-context inference without demonstrations, privileged inputs, or test-time gradient updates.

Across PegInsert, GearMesh, and NutThread, CoRMA provides a reusable adaptation interface within a related assembly family and retains higher verified real success than FORGE baselines under the same Marvin deployment interface. 
The results suggest that semantic contact inference from force history is a promising direction for robust sim-to-real assembly. 
Future work should evaluate held-out tasks, train a single multi-task teacher and shared adapter, perform Real2Sim calibration, and improve deployment-time force estimation.

% \section*{Acknowledgments}

%% Use plainnat to work nicely with natbib. 

% \bibliographystyle{plainnat}
\newpage
\bibliography{references}

\newpage

\appendix

\section{Additional Experimental Results}
\label{app:additional_results}

This appendix provides supplementary results for the Stage~2 adapter and real-world force analysis. 
Appendix~\ref{app:ablation_experiments} reports ablation studies for the adapter architecture and training objective.
Appendix~\ref{app:real_force_regimes} shows real wrench evidence motivating the four force-regime labels used for contrastive learning.
Appendix~\ref{app:stage2_diagnostics} provides additional diagnostics showing that the Stage~2 adapter learns a structured and predictive contact representation.

\section{Stage-2 Ablation Experiments}
\label{app:ablation_experiments}

Table~\ref{tab:stage2_param_ablation} studies the effect of adapter capacity and auxiliary loss weights on 6D Privileged $Z$ prediction. 
The large causal Transformer with a weak contrastive weight, NCE $\lambda=0.01$, gives the best overall validation performance.
This supports the use of a high-capacity temporal encoder with contrastive learning as a regularizer rather than as a dominant objective.

\begin{table}[t]
\centering
\caption{
\textbf{Stage-2 adapter hyperparameter ablation.}
All entries evaluate 6D Privileged $Z$ prediction on the validation set.
Higher $R^2$, Pearson correlation, and cosine similarity are better; lower MSE is better.
Failed or missing runs are omitted.
}
\vspace{4px}
\label{tab:stage2_param_ablation}

\begin{tabular}{llcccc}
\toprule
\textbf{ID} & \textbf{Configuration} & $\mathbf{R^2}$ & \textbf{Pearson} & \textbf{Cos-Sim} & \textbf{MSE} \\
\midrule
A2 & MSE only & 0.7453 & 0.8572 & 0.7815 & 0.2547 \\
A3 & NCE $\lambda=0.5$ & 0.6799 & 0.8135 & 0.7396 & 0.3201 \\
A4 & MSE + NCE + smooth $\lambda=0.05$ & 0.6782 & 0.8145 & 0.7414 & 0.3218 \\
B1 & NCE temperature $=0.5$ & 0.7413 & 0.8546 & 0.7777 & 0.2587 \\
B2 & NCE temperature $=0.05$ & 0.7131 & 0.8360 & 0.7612 & 0.2869 \\
C1 & Small Transformer, $d=64$, 2 layers & 0.6640 & 0.8016 & 0.7314 & 0.3360 \\
C2 & Large Transformer, $d=256$, 4 layers & 0.8148 & 0.9000 & 0.8335 & 0.1852 \\
C3 & Dropout $=0.1$ & 0.6792 & 0.8170 & 0.7405 & 0.3208 \\
D2 & Large + NCE $\lambda=0.01$ & \textbf{0.9044} & \textbf{0.9502} & \textbf{0.9136} & \textbf{0.0956} \\
D3 & Large + NCE $\lambda=0.05$ & 0.8115 & 0.8978 & 0.8301 & 0.1885 \\
D4 & Large + NCE $\lambda=0.2$ & 0.7886 & 0.8843 & 0.8134 & 0.2114 \\
D6 & Large + NCE $\lambda=0.5$ & 0.7639 & 0.8693 & 0.7945 & 0.2361 \\
D7 & Large + NCE $\lambda=0.1$, temp $=0.05$ & 0.8492 & 0.9196 & 0.8629 & 0.1508 \\
D8 & Large + NCE $\lambda=0.1$, temp $=0.5$ & 0.8556 & 0.9232 & 0.8684 & 0.1444 \\
D9 & Large + dropout $=0.1$ & 0.7662 & 0.8733 & 0.8009 & 0.2338 \\
E1 & Large + NCE $\lambda=0.01$ + texture $\lambda=0.05$ & 0.8380 & 0.9132 & 0.8507 & 0.1620 \\
E2 & Large + NCE $\lambda=0.01$ + texture $\lambda=0.1$ & 0.8025 & 0.8926 & 0.8219 & 0.1975 \\
E3 & Large + texture $\lambda=0.1$, no NCE & 0.8112 & 0.8976 & 0.8278 & 0.1888 \\
\bottomrule
\end{tabular}

\end{table}

Table~\ref{tab:stage2_structure_ablation} isolates the architectural and objective-level contributions.
Replacing the causal Transformer with a Conv1D RMA-style adapter causes a large performance drop across all task pairs.
The contrastive objective gives a smaller but consistent aggregate gain over MSE-only training, indicating that force-regime structure improves representation quality without sacrificing latent prediction accuracy.

\begin{table}[t]
\centering
\caption{
\textbf{Stage-2 structure ablation across task pairs.}
CoRMA uses a causal Transformer adapter with semantic regression and force-regime InfoNCE.
The MSE-only variant removes the contrastive objective, while RMA-Conv replaces the causal Transformer with a Conv1D RMA-style adapter.
Mean values are computed across the three task pairs.
}
\vspace{4px}
\label{tab:stage2_structure_ablation}
\begin{tabular}{llcccc}
\toprule
\textbf{Method} & \textbf{Task Pair} & $\mathbf{R^2}$ & \textbf{Pearson} & \textbf{Cos-Sim} & \textbf{MSE} \\
\midrule
CoRMA & GearMesh--NutThread & 0.9025 & 0.9497 & 0.9455 & 0.0975 \\
CoRMA & GearMesh--PegInsert & 0.8555 & 0.9234 & 0.8915 & 0.1445 \\
CoRMA & NutThread--PegInsert & 0.8797 & 0.9375 & 0.9275 & 0.1203 \\
\midrule
Transformer-MSE & GearMesh--NutThread & 0.9053 & 0.9512 & 0.9470 & 0.0947 \\
Transformer-MSE & GearMesh--PegInsert & 0.8507 & 0.9210 & 0.8896 & 0.1493 \\
Transformer-MSE & NutThread--PegInsert & 0.8505 & 0.9217 & 0.9088 & 0.1495 \\
\midrule
RMA-Conv & GearMesh--NutThread & 0.5220 & 0.7184 & 0.7172 & 0.4780 \\
RMA-Conv & GearMesh--PegInsert & 0.4872 & 0.6820 & 0.6776 & 0.5128 \\
RMA-Conv & NutThread--PegInsert & 0.2915 & 0.5229 & 0.4972 & 0.7084 \\
\midrule
\textbf{CoRMA Mean} & All pairs & \textbf{0.8792} & \textbf{0.9369} & \textbf{0.9215} & \textbf{0.1208} \\
Transformer-MSE Mean & All pairs & 0.8688 & 0.9313 & 0.9151 & 0.1312 \\
RMA-Conv Mean & All pairs & 0.4336 & 0.6411 & 0.6307 & 0.5664 \\
\bottomrule
\end{tabular}
\end{table}

\section{Real-Force Motivation for Contact Regime Labels}
\label{app:real_force_regimes}

\begin{figure*}[t]
    \centering
    \includegraphics[width=\linewidth]{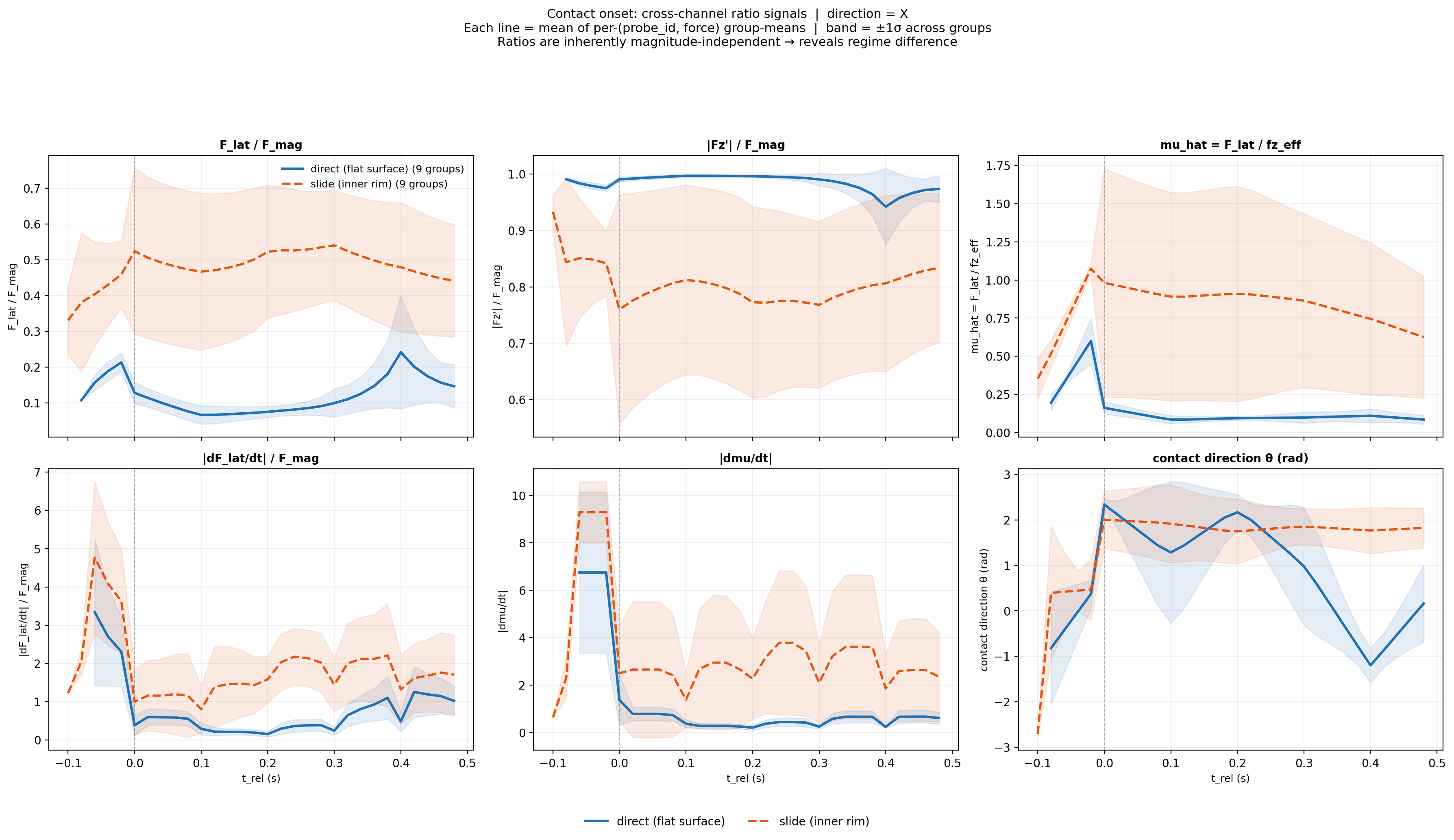}
    \caption{
    \textbf{Real force-regime evidence for positive $X$ contact direction.}
    Wrench-derived features are aligned at contact onset. 
    Direct flat-surface contact and inner-rim sliding show different lateral-force ratios, force derivatives, and contact-direction signatures, motivating the coarse contact regimes used for contrastive supervision.
    }
    \label{fig:onset_summary_X}
\end{figure*}

\begin{figure*}[t]
    \centering
    \includegraphics[width=\linewidth]{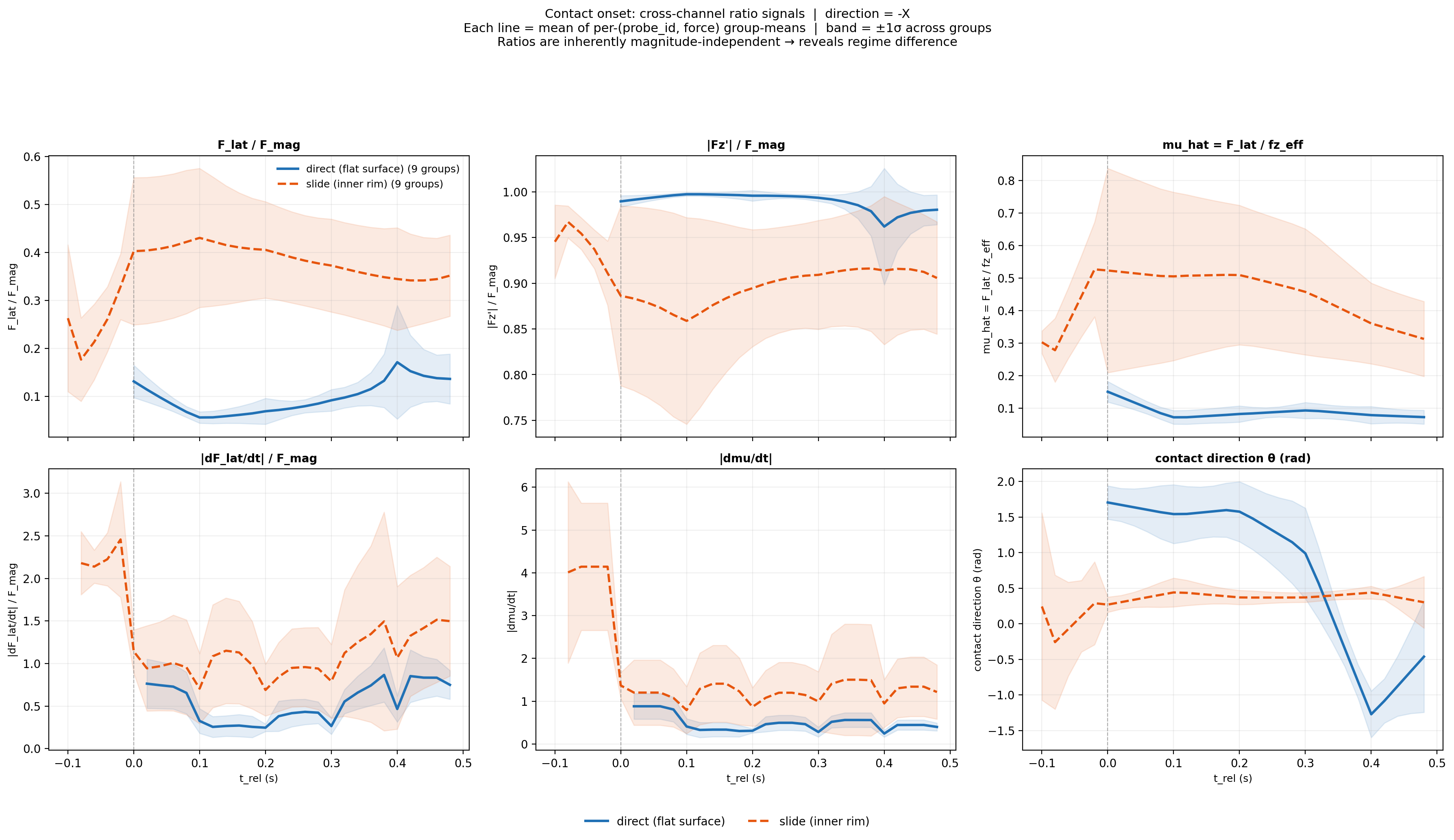}
    \caption{
    \textbf{Real force-regime evidence for negative $X$ contact direction.}
    The sharp change near contact onset indicates the transition from free motion to first contact, while the sustained post-onset force texture differentiates guided sliding from direct contact.
    }
    \label{fig:onset_summary_negX}
\end{figure*}

\begin{figure*}[t]
    \centering
    \includegraphics[width=\linewidth]{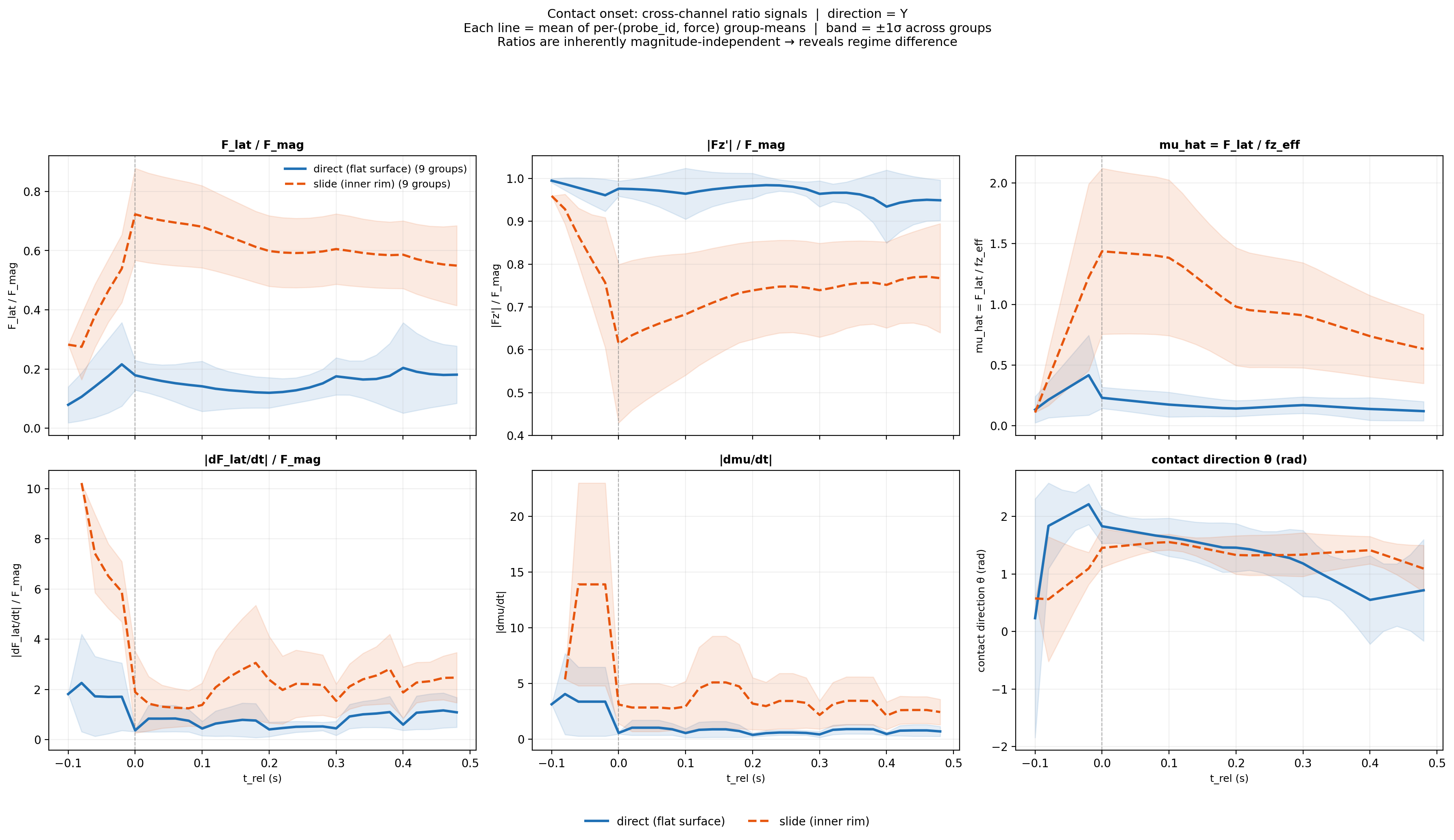}
    \caption{
    \textbf{Real force-regime evidence for positive $Y$ contact direction.}
    The lateral-force ratio, normalized force derivative, and contact-direction angle provide deployable wrench cues for separating first contact, guided sliding, and jam-like interaction.
    }
    \label{fig:onset_summary_Y}
\end{figure*}

\begin{figure*}[t]
    \centering
    \includegraphics[width=\linewidth]{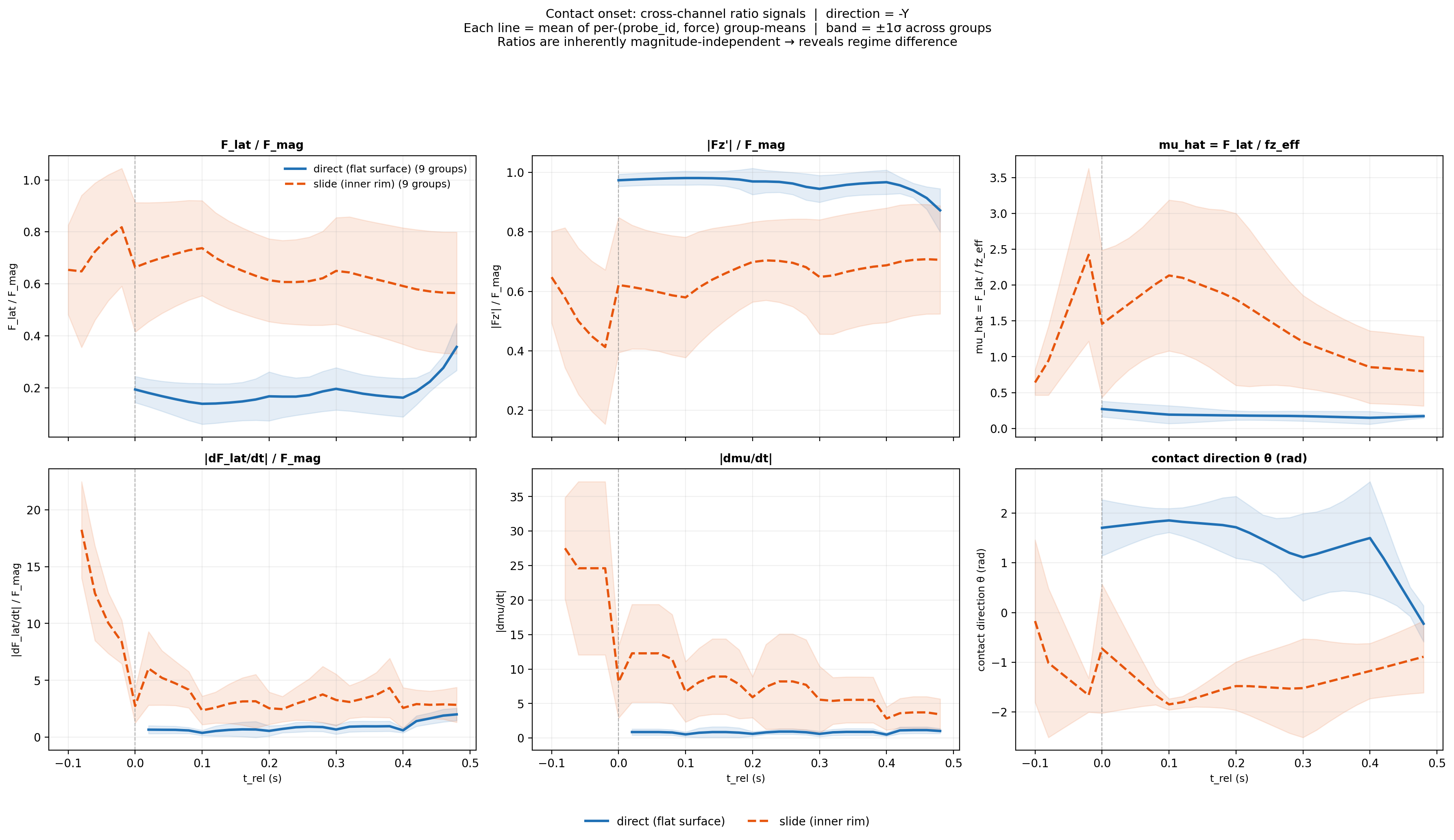}
    \caption{
    \textbf{Real force-regime evidence for negative $Y$ contact direction.}
    Across probe groups, inner-rim sliding exhibits sustained lateral-force structure and directionally consistent contact signatures, supporting the use of coarse force-regime labels in Stage~2.
    }
    \label{fig:onset_summary_negY}
\end{figure*}

The Stage~2 contrastive objective uses four weak force-regime labels:
\[
\texttt{free},\quad
\texttt{first\_contact},\quad
\texttt{guided\_slide},\quad
\texttt{jam}.
\]
These labels are not intended to provide exact physical state estimation. 
Instead, they define coarse positive and negative pairs for contrastive representation learning from deployable force history.
Figs.~\ref{fig:onset_summary_X}--\ref{fig:onset_summary_negY} show real wrench-derived features aligned at contact onset for different contact directions.
Across trials, the transition from free motion to first contact produces a sharp change in lateral force ratio and lateral-force derivative near $t=0$.
Sliding contact along the inner rim produces sustained force texture and a distinct contact-direction signature compared with direct flat-surface contact.
These repeated patterns motivate the use of force-regime labels as weak semantic supervision for CoRMA.

\section{Stage-2 Adapter Diagnostics}
\label{app:stage2_diagnostics}

This section provides supplementary diagnostics for the Stage~2 CoRMA adapter.
The main paper reports quantitative validation in Table~\ref{tab:adapter_summary}; here we visualize how the supervised semantic output $\hat z_t$ and the auxiliary contrastive embedding $u_t$ organize validation histories.
These diagnostics support the two-head adapter design: the semantic head should preserve numerical fidelity to the simulator-derived 6D Privileged $Z$, while the contrastive head should organize histories by coarse force-regime semantics.
All plots in this section are diagnostic and do not imply fully task-invariant representation learning.

\subsection{Contrastive Regime Organization}
\label{app:contrastive_regime_organization}

\begin{figure*}[t]
  \centering
  \includegraphics[width=\linewidth]{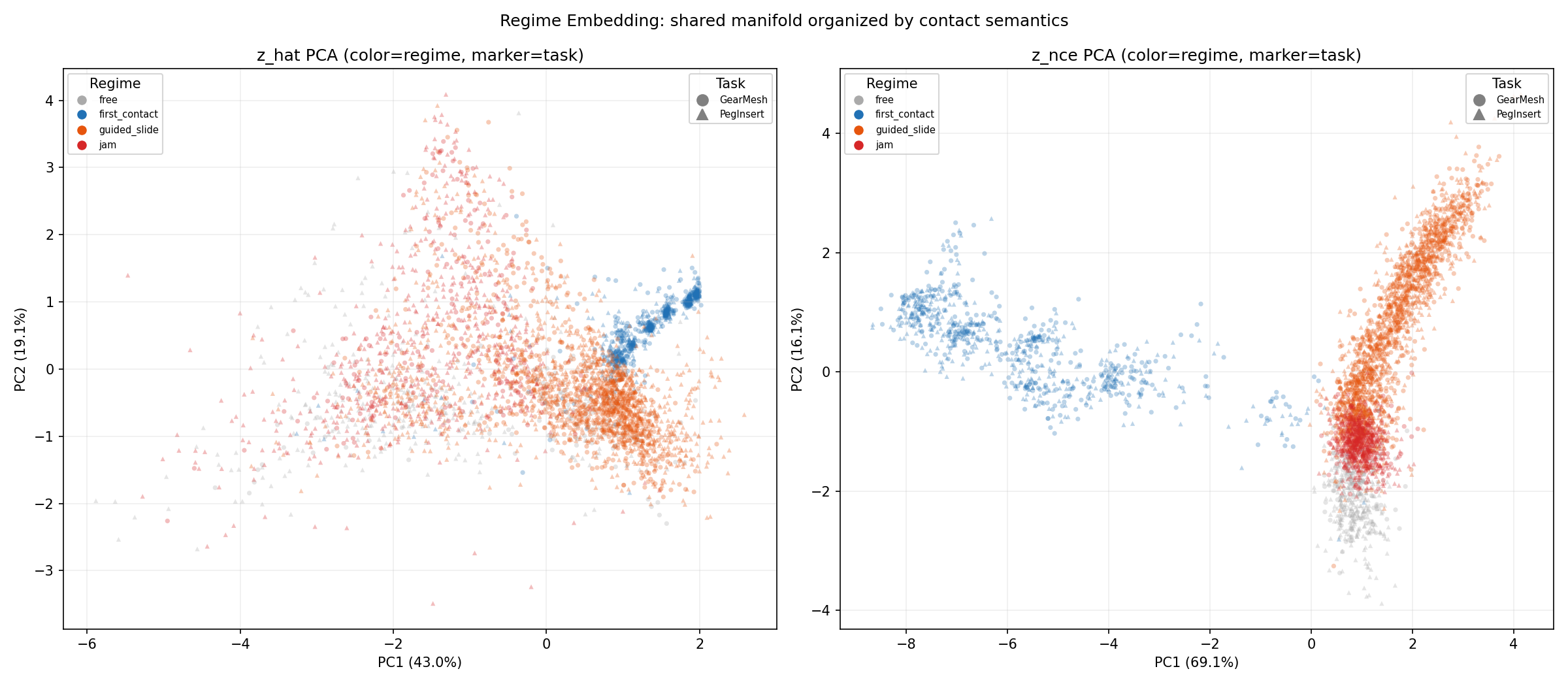}
  \caption{
  \textbf{GearMesh--PegInsert adapter embedding diagnostic.}
  Left: PCA of the predicted semantic context $\hat z_t$.
  Right: PCA of the auxiliary contrastive embedding $u_t$.
  Colors indicate coarse force regimes and markers indicate task identity.
  Compared with $\hat z_t$, the contrastive embedding shows clearer regime-level organization, supporting the use of force-regime positives and negatives for InfoNCE.
  }
  \label{fig:gearpeg_regime_embedding}
\end{figure*}

\begin{figure*}[t]
  \centering
  \includegraphics[width=\linewidth]{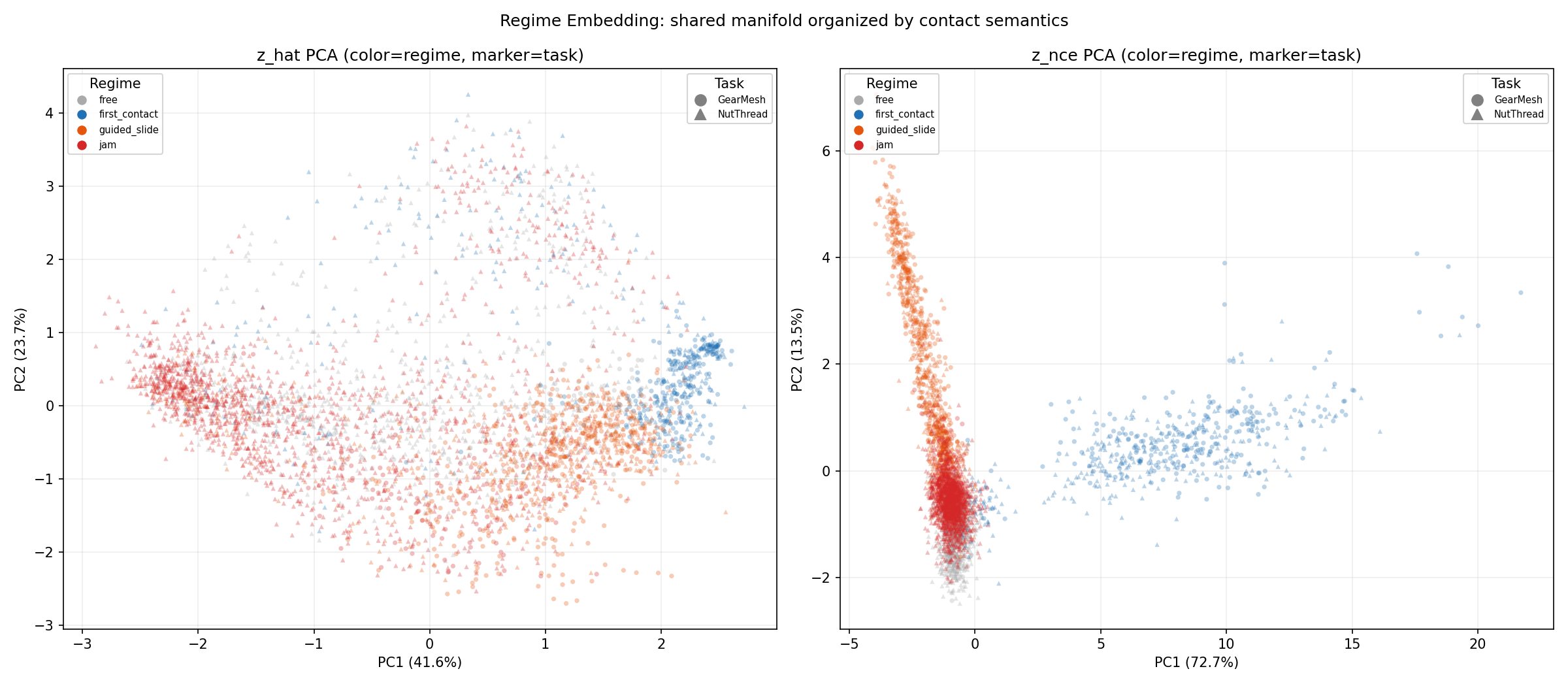}
  \caption{
  \textbf{GearMesh--NutThread adapter embedding diagnostic.}
  Left: PCA of $\hat z_t$.
  Right: PCA of $u_t$.
  The auxiliary contrastive embedding provides clearer force-regime organization than the semantic regression output alone.
  This visualization supports the contrastive head as a representation-structuring objective, not as a replacement for supervised Privileged $Z$ regression.
  }
  \label{fig:gearnut_regime_embedding}
\end{figure*}

\begin{figure*}[t]
  \centering
  \includegraphics[width=\linewidth]{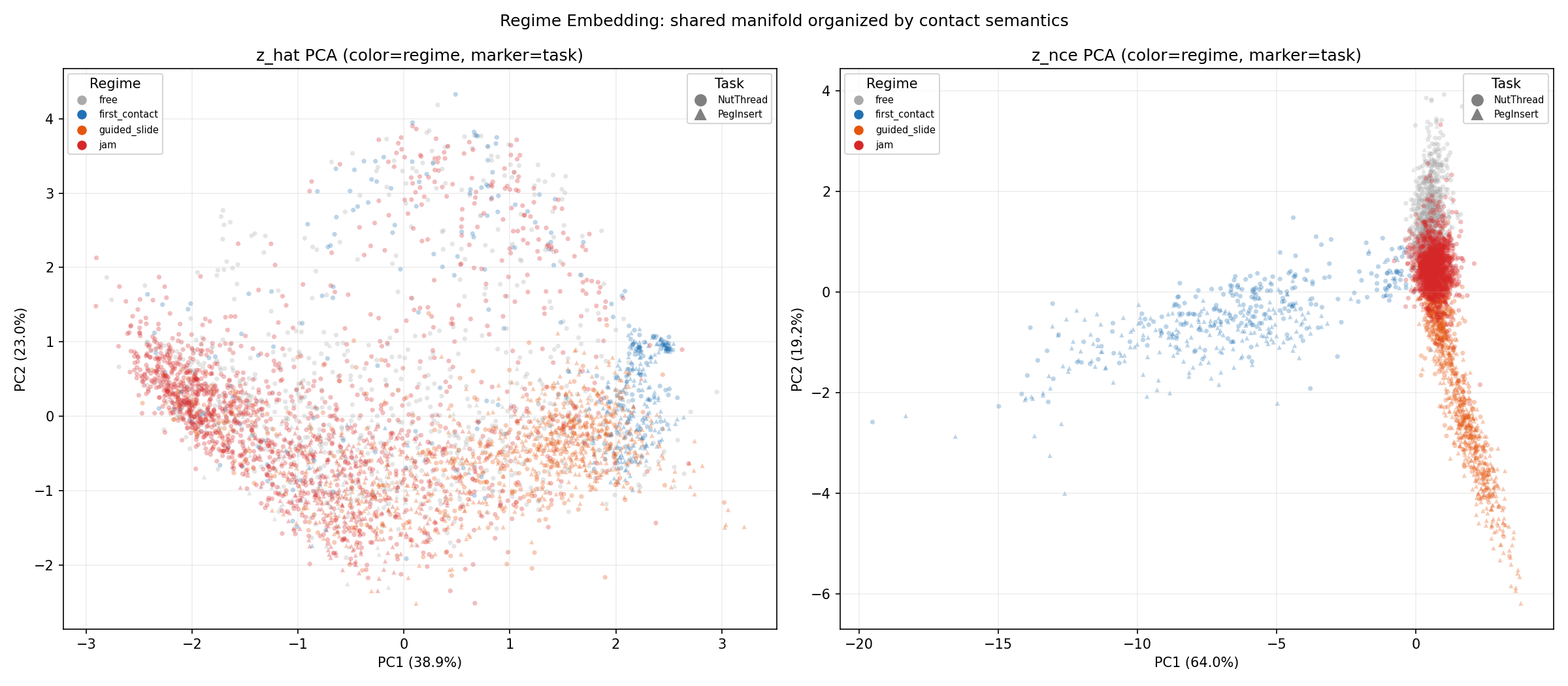}
  \caption{
  \textbf{NutThread--PegInsert adapter embedding diagnostic.}
  Left: PCA of $\hat z_t$.
  Right: PCA of $u_t$.
  The contrastive space separates coarse contact regimes more clearly than the regression output alone, indicating that force-regime positives and negatives provide useful semantic structure.
  Task-specific structure may still remain, so this should be interpreted as qualitative diagnostic evidence rather than proof of complete task invariance.
  }
  \label{fig:nutpeg_regime_embedding}
\end{figure*}

The Stage~2 InfoNCE objective defines positives as histories with the same coarse force-regime label and negatives as histories with different labels, regardless of task identity.
Therefore, if the contrastive objective is working as intended, the auxiliary embedding $u_t$ should show clearer regime-level organization than the regression output $\hat z_t$ alone.
Figs.~\ref{fig:gearpeg_regime_embedding}--\ref{fig:nutpeg_regime_embedding} compare these two spaces for the three pairwise adapters.
In each figure, the left panel shows PCA of $\hat z_t$, which is trained for numerical prediction of the 6D Privileged $Z$; the right panel shows PCA of $u_t$, which is trained only by the force-regime contrastive head.
Across task pairs, $u_t$ exhibits more separated force-regime structure, especially for first-contact and guided-sliding histories.
This supports the use of force-regime InfoNCE as a semantic regularizer, while the remaining overlap and task-specific structure indicate that the representation is not fully task-invariant.

\subsection{Semantic Context Prediction Fidelity}
\label{app:semantic_prediction_fidelity}

\begin{figure*}[t]
  \centering
  \includegraphics[width=\textwidth,trim=20 20 20 35,clip]{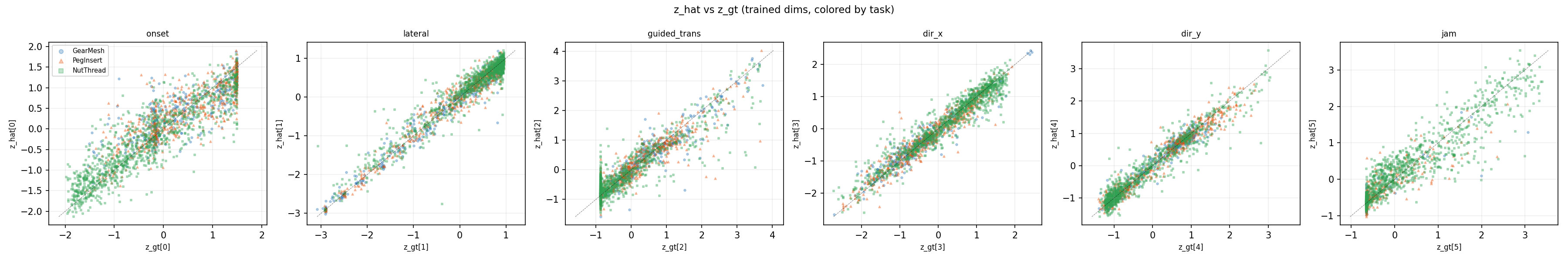}
  \caption{
  \textbf{GearMesh--PegInsert semantic context prediction.}
  Predicted $\hat z_t$ is plotted against oracle Privileged $Z$ for the six semantic dimensions.
  The diagonal trend indicates that the adapter preserves numerical fidelity to the simulator-derived contact context.
  }
  \label{fig:gearpeg_zhat_zgt}
\end{figure*}

\begin{figure*}[t]
  \centering
  \includegraphics[width=\textwidth,trim=20 20 20 35,clip]{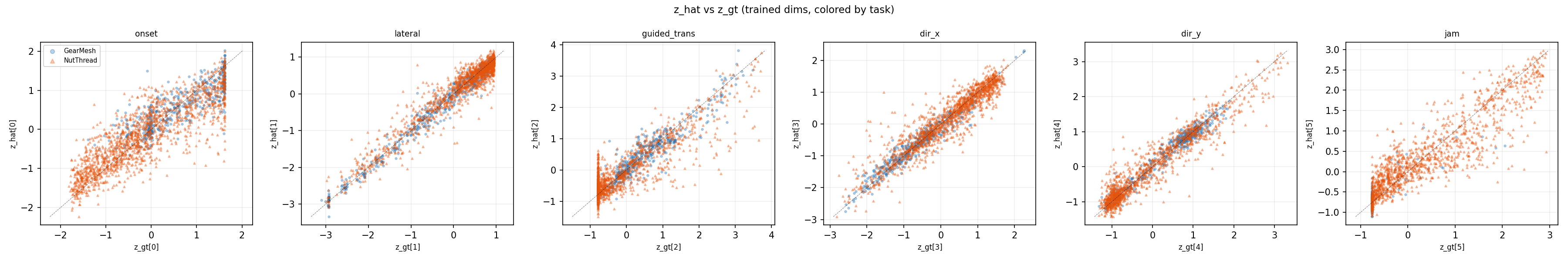}
  \caption{
  \textbf{GearMesh--NutThread semantic context prediction.}
  Predicted $\hat z_t$ follows oracle $z_t$ across the trained semantic dimensions, supporting the use of $\hat z_t$ as the deployable context injected into the policy.
  }
  \label{fig:gearnut_zhat_zgt}
\end{figure*}

\begin{figure*}[t]
  \centering
  \includegraphics[width=\textwidth,trim=20 20 20 35,clip]{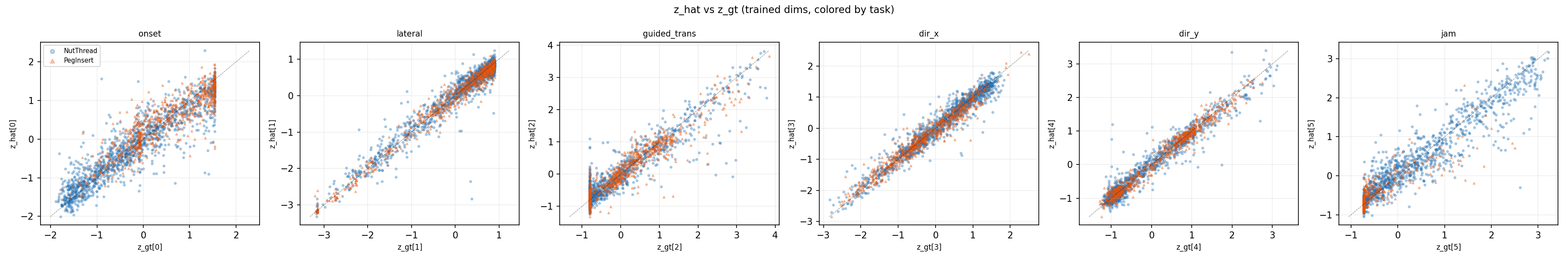}
  \caption{
  \textbf{NutThread--PegInsert semantic context prediction.}
  The adapter prediction $\hat z_t$ remains aligned with oracle Privileged $Z$ across contact semantics, showing that the contrastive auxiliary objective does not replace or collapse the supervised semantic regression target.
  }
  \label{fig:nutpeg_zhat_zgt}
\end{figure*}

The policy receives $\hat z_t$, not the auxiliary contrastive embedding $u_t$, during Stage~3 fine-tuning and real deployment.
Therefore, the contrastive head is useful only if the semantic head still predicts the simulator-derived Privileged $Z$ accurately.
Figs.~\ref{fig:gearpeg_zhat_zgt}--\ref{fig:nutpeg_zhat_zgt} compare predicted $\hat z_t$ against oracle $z_t$ for the six trained semantic dimensions across the three pairwise adapters.
The points concentrate around the diagonal across contact onset, lateral engagement, guided transition, contact direction, and jam tendency, indicating that the adapter learns a deployable approximation of the privileged semantic context rather than a constant latent.
Together with the contrastive embedding diagnostics above, these plots support the intended division of labor in CoRMA: supervised regression preserves the control-relevant 6D context, while InfoNCE shapes an auxiliary representation by contact regime.

\subsection{Force-Regime Separability from Predicted Context}
\label{app:regime_separability}

\begin{figure*}[t]
  \centering
  \begin{subfigure}[t]{0.32\textwidth}
    \centering
    \includegraphics[width=\linewidth]{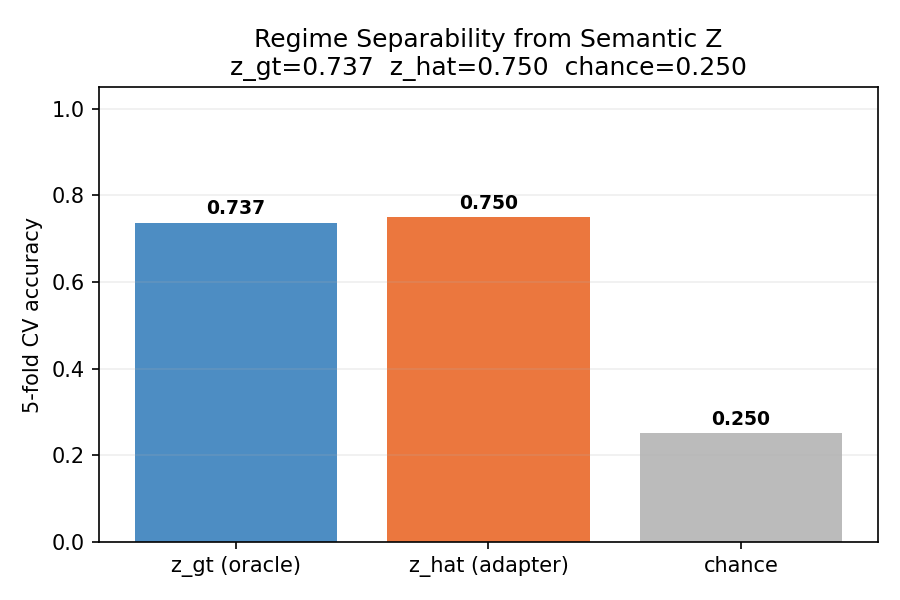}
    \caption{GearMesh--PegInsert}
    \label{fig:gearpeg_regime_separability}
  \end{subfigure}
  \hfill
  \begin{subfigure}[t]{0.32\textwidth}
    \centering
    \includegraphics[width=\linewidth]{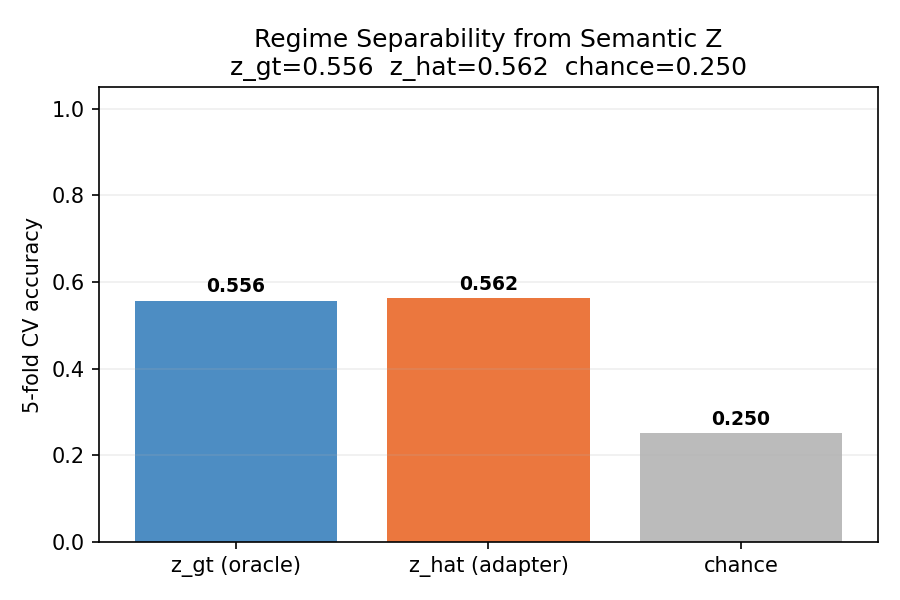}
    \caption{GearMesh--NutThread}
    \label{fig:gearnut_regime_separability}
  \end{subfigure}
  \hfill
  \begin{subfigure}[t]{0.32\textwidth}
    \centering
    \includegraphics[width=\linewidth]{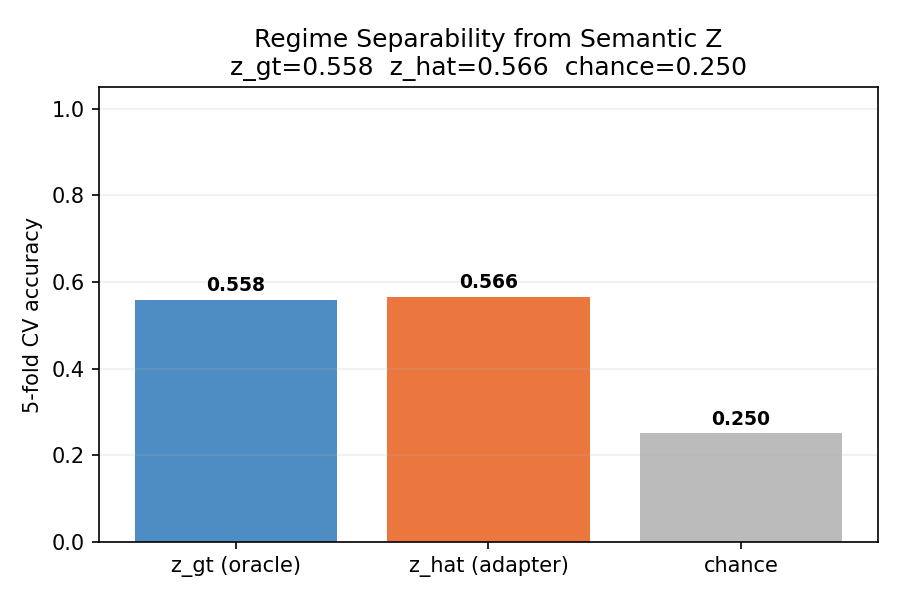}
    \caption{NutThread--PegInsert}
    \label{fig:nutpeg_regime_separability}
  \end{subfigure}
  \caption{
  \textbf{Force-regime separability from semantic context.}
  For each pairwise adapter, a simple probe predicts coarse force-regime labels from oracle $z_t$ and predicted $\hat z_t$.
  The predicted context remains above chance and close to the oracle context across task pairs, indicating that $\hat z_t$ preserves regime-relevant semantic information.
  This diagnostic supports the use of $\hat z_t$ as a deployable contact context, but does not by itself prove downstream causal improvement or full task-invariant generalization.
  }
  \label{fig:pairwise_regime_separability}
\end{figure*}

We further evaluate whether the predicted semantic context $\hat z_t$ preserves the coarse force-regime information used to define contrastive positives and negatives.
For each pairwise adapter, we train a simple probe to predict the four force-regime labels from either oracle $z_t$ or predicted $\hat z_t$.
If $\hat z_t$ retains the semantic structure of Privileged $Z$, its regime-prediction accuracy should be close to the oracle $z_t$ probe and above the chance level of $25\%$.
Fig.~\ref{fig:pairwise_regime_separability} shows that the predicted context remains regime-informative across task pairs.
This diagnostic complements the PCA visualizations: the contrastive head structures the auxiliary embedding $u_t$, while the deployed context $\hat z_t$ still carries regime-relevant semantic information for policy conditioning.

\end{document}